\PassOptionsToPackage{prologue,dvipsnames}{xcolor}
\documentclass[10pt,twocolumn,letterpaper]{article}

\usepackage[pagenumbers]{iccv} % To force page numbers, e.g. for an arXiv version

\definecolor{iccvblue}{rgb}{0.21,0.49,0.74}
\usepackage[pagebackref,breaklinks,colorlinks,allcolors=iccvblue]{hyperref}

\usepackage{multirow}
\PassOptionsToPackage{table}{xcolor}
\usepackage[table]{xcolor}
\usepackage{colortbl}
\usepackage{algorithm}
\usepackage{algpseudocode}
\usepackage{url}
\usepackage[dvipsnames]{xcolor}
\usepackage{graphicx}

\usepackage{pifont}
\newcommand{\cmark}{\textcolor{ForestGreen}{\ding{51}}} 
\newcommand{\xmark}{\textcolor{red}{\ding{55}}} 

\DeclareMathOperator*{\argmin}{arg\,min}

\def\x{{\boldsymbol{x}}}
\def\y{{\boldsymbol{y}}}
\def\z{{\boldsymbol{z}}}

\def\X{{\boldsymbol{X}}}
\def\Y{{\boldsymbol{Y}}}
\def\Z{{\boldsymbol{Z}}}

\def\Ac{{\mathcal{A}}}

\newcommand{\addorange}[1] {\textcolor{BurntOrange}{#1}}
\newcommand{\addblue}[1] {\textcolor{CornflowerBlue}{#1}}

\def\paperID{8243} % *** Enter the Paper ID here
\def\confName{ICCV}
\def\confYear{2025}

\title{VISION-XL: High Definition Video Inverse Problem Solver using Latent Image Diffusion Models}

\author{Taesung Kwon\\
KAIST\\
{\tt\small star.kwon@kaist.ac.kr}
\and
Jong Chul Ye\\
KAIST\\
{\tt\small jong.ye@kaist.ac.kr}
}

\begin{document}

\twocolumn[{%
\renewcommand\twocolumn[1][]{#1}%
\maketitle
\begin{center}
    \centering
    \captionsetup{type=figure}
    \vspace{-0.3cm}
    \includegraphics[width=\textwidth]{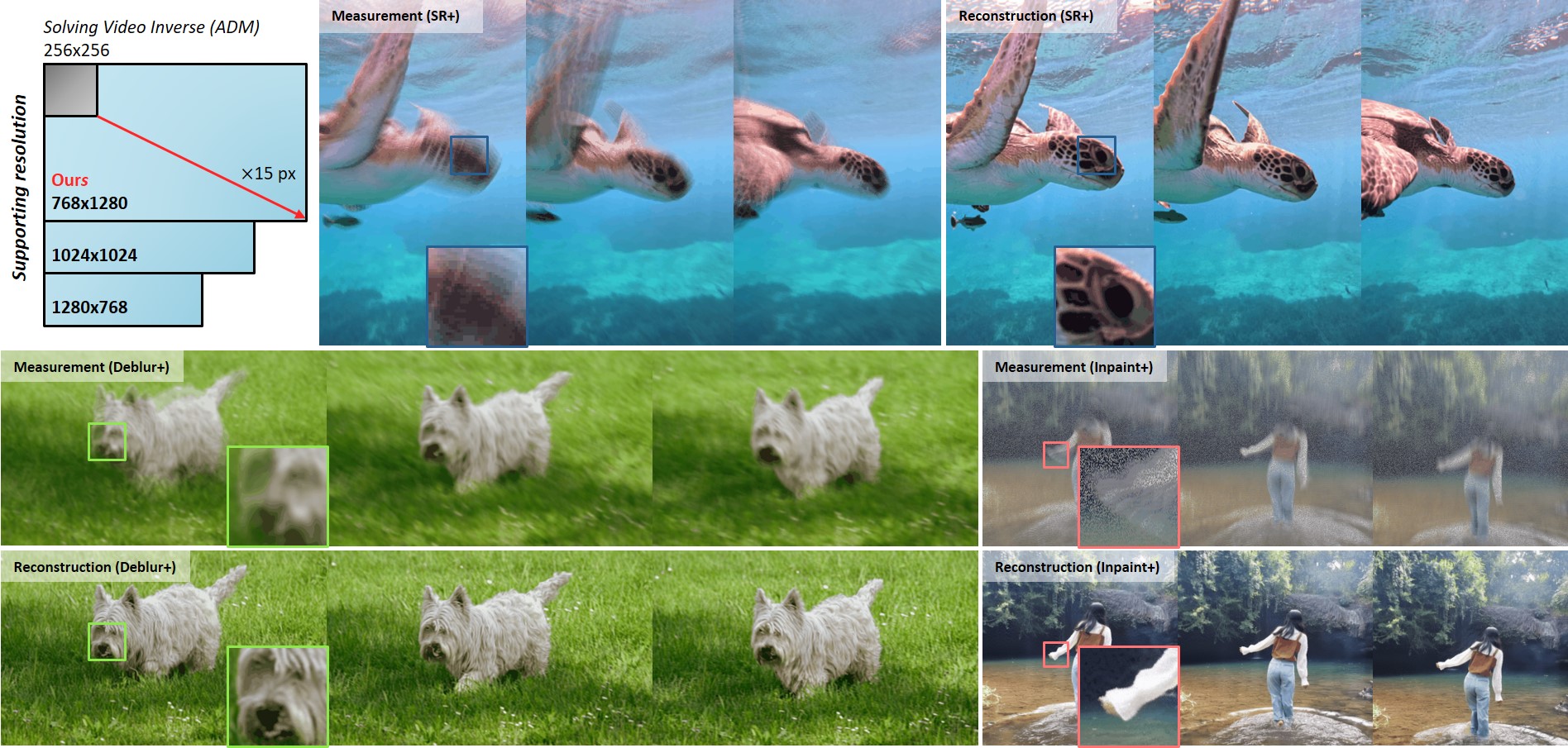}
    \vspace{-0.6cm}
    \captionof{figure}{Representative video reconstruction by VISION-XL: SR+ (frame averaging with $\times$4 super-resolution, top), Deblur+ (frame averaging with deblurring, $\sigma$=3.0, bottom-left), and Inpaint+ (frame averaging with 50\% random inpainting, bottom-right).}
    % \vspace{-0.2cm}
\end{center}%
}]

\begin{abstract}
In this paper, we propose a novel framework for solving high-definition video inverse problems using latent image diffusion models. 
Building on recent advancements in spatio-temporal optimization for video inverse problems using image diffusion models,
 our approach leverages latent-space diffusion models to achieve enhanced video quality and resolution.
To address the high computational demands of processing high-resolution frames, we introduce a pseudo-batch consistent sampling strategy, allowing efficient operation on a single GPU.
Additionally, to improve temporal consistency, we present pseudo-batch inversion, an initialization technique that incorporates informative latents from the measurement.
By integrating with SDXL, our framework achieves state-of-the-art video reconstruction across a wide range of spatio-temporal inverse problems, including complex combinations of frame averaging and various spatial degradations, such as deblurring, super-resolution, and inpainting.
Unlike previous methods, our approach supports multiple aspect ratios (landscape, vertical, and square) and delivers HD-resolution reconstructions (exceeding 1280×720) in {under 6 seconds per frame} on a single NVIDIA 4090 GPU.
Project page: \small \textsf{\url{https://vision-xl.github.io/}}.
\end{abstract}

\vspace{-0.3cm}

\section{Introduction}
\label{sec:intro}

Diffusion models~\cite{ho2020denoising, song2020score, dhariwal2021diffusion, song2021denoising, rombach2022high, karras2022elucidating, podell2023sdxl} have set a new benchmark in generative modeling, enabling the generation of high-quality samples. 
These models have become the foundation for advancements in various fields, such as controllable image editing~\cite{zhang2023adding}, image personalization~\cite{gal2022image}, synthetic data augmentation~\cite{stockl2023evaluating}, and even reconstructing images from brain signals~\cite{takagi2023high, li2024visual}.

Furthermore, diffusion model-based inverse problem solvers (DIS)~\cite{kawar2022denoising, song2023pseudoinverseguided,  chung2023diffusion, wang2023zeroshot, chung2024decomposed, xiao2024dreamclean} address a variety of image restoration tasks, such as deblurring, super-resolution, inpainting, colorization, compressed sensing, and so on.
A key feature of DIS is its plug-and-play capability, allowing diffusion models to be applied flexibly across different inverse problems without requiring task-specific training or fine-tuning.

Recently, several extensions~\cite{kwon2024solving, yeh2024diffir2vr, daras2024warped} from the DIS have been proposed to solve video inverse problems using image diffusion models.
Naive application of image diffusion models to videos may break temporal consistency.
To address this problem, these methods preserve temporal consistency by utilizing a batch-consistent sampling strategy~\cite{kwon2024solving} and applying optical flow guidance to warp either latent representations~\cite{yeh2024diffir2vr} or the noise prior~\cite{daras2024warped}. 

Although these innovative approaches enable powerful image generative models~\cite{dhariwal2021diffusion, rombach2022high, podell2023sdxl} to solve video inverse problems with significantly reduced computational requirements, there is still room for improvement in these methods.
Optical flow-based methods~\cite{yeh2024diffir2vr, daras2024warped} have reported a key limitation: their performance is highly dependent on the accuracy of the optical flow estimation module~\cite{teed2020raft, xu2022gmflow}. 
This dependency becomes problematic when extreme degradations complicate the estimation process, restricting their applicability to a wider range of restoration tasks.
Additionally, these methods require task-specific restoration modules~\cite{yeh2024diffir2vr} or fine-tuning of the diffusion model~\cite{daras2024warped}.
In this perspective, batch consistent sampling strategy~\cite{kwon2024solving} successfully addressed various spatio-temporal degradations without requiring task-specific training or fine-tuning.
%Furthermore, we demonstrate that a direct extension of SVI~\cite{kwon2024solving} to latent diffusion models fails to maintain temporal consistency, as shown in Fig.~\ref{fig:initialization}.
However, we empirically found that extending SVI~\cite{kwon2024solving} to latent diffusion models {leads to unsatisfactory reconstruction, particularly in terms of FVD~\cite{unterthiner2019fvd},} as shown in Table~\ref{table:initialization}, despite the fact that most modern latent diffusion models are essential for scaling up to large-size video inverse problems.
% However, the reconstruction resolution is limited to 256$\times$256, as it utilizes the unconditional {\em pixel}-space diffusion model provided by ADM~\cite{dhariwal2021diffusion}.

\begin{table}[!t]
\centering
% \vspace{-0.5cm}
\resizebox{\columnwidth}{!}{
\begin{tabular}[t]{cccc}
\hline
Methods&Spatio-temporal degradation&Latent space\\
\hline
Optical flow-based~\cite{yeh2024diffir2vr,daras2024warped}&\xmark&\cmark\\
SVI~\cite{kwon2024solving}&\cmark&\xmark\\
Ours&\cmark&\cmark\\
\hline
\end{tabular}
}
\vspace{-0.25cm}
\caption{{Comparison of image diffusion-based video inverse problem solvers. Unlike other methods, our approach leverages latent image diffusion models to address spatio-temporal degradation, enabling more effective restoration.}}
\label{table:comparison}
\vspace{-0.5cm}
\end{table}%

To overcome this limitations, here we propose a novel framework for solving high-definition video inverse problems using latent image diffusion models.
To address the high computational demands of batch processing (e.g., 16-frame batch processing used in~\cite{kwon2024solving}) with high-resolution latent diffusion models~\cite{podell2023sdxl}, we introduce  {\em  pseudo-batch consistent sampling}.
This strategy enables multi-frame video processing while requiring only the memory needed for a single frame, making it feasible on a single GPU.
Furthermore, we propose {\em pseudo-batch inversion}, which initializes the process with informative latents derived from the measurement.
This initialization enhances temporal consistency and improves the efficiency of solving spatio-temporal inverse problems as shown in Table.~\ref{table:initialization}.

By integrating these components, our framework achieves state-of-the-art video reconstruction performance using SDXL~\cite{podell2023sdxl}. 
We name the method integrating these components as VISION-XL, short for \textbf{V}ideo \textbf{I}nverse-problem \textbf{S}olver using latent diffus\textbf{ION} models (with stable diffusion \textbf{XL}).
It supports various aspect ratios, including landscape, vertical, and square formats. Thanks to its efficiency, our framework can reconstruct 1280×768 (exceeding HD resolution) videos in under 6 seconds per frame on a single NVIDIA 4090 GPU. Our contribution can be summarized as follows:
\begin{itemize}
    \item {We propose a high-definition video inverse problem solver integrated with SDXL, supporting multiple aspect ratios and achieving state-of-the-art reconstruction.}
    \item {We introduce a novel pseudo-batch consistent sampling and inversion strategy for efficient and effective video reconstruction across diverse inverse problems.}
\end{itemize}

\begin{table}[t]
\centering
% \vspace{-0.5cm}
    \resizebox{0.9\linewidth}{!}{
        \begin{tabular}{ccccc}
        \toprule
        Initialization & FVD $\downarrow$ & LPIPS $\downarrow$ & PSNR $\uparrow$ \\
        \cmidrule(lr){1-1}\cmidrule(lr){2-4}
        Random noise  & 1047 & 0.251  & 29.43  \\
        Batch synchronized noise (SVI~\cite{kwon2024solving}) & 707.7 & 0.248 & 30.10 \\
        \cmidrule(lr){1-4}
        Pseudo-batch inversion (Ours) & \textbf{184.8} & \textbf{0.236} & \textbf{30.74} \\ 
        \bottomrule
        \end{tabular}
    }
    \vspace{-0.2cm}
\caption{{Impact of our initialization method on SR+ video restoration using SDXL~\cite{podell2023sdxl}. Our method significantly improves performance, reducing FVD~\cite{unterthiner2019fvd} by more than 3$\times$.}}
\vspace{-0.5cm}
\label{table:initialization}
\end{table}

% \begin{figure}[!t]
%   \centering
%     \centerline{{\includegraphics[width=0.9\linewidth]{figures/initialization.jpg}}}
%     \vspace{-0.2cm}
%     \caption{\add{Effect of initialization in Deblur+ restoration. Batch-synchronized noise sampling (SVI~\cite{kwon2024solving}) often leads to intensity fluctuations, whereas our method significantly enhances performance, particularly in FVD~\cite{unterthiner2019fvd}.}}
%     \label{fig:initialization}
%     \vspace{-0.5cm}
% \end{figure}

\section{Related Work}
\label{sec:related}

\noindent\textbf{Diffusion model-based inverse problem solvers (DIS).}
Diffusion models~\cite{ho2020denoising, song2020score, song2021denoising} attempt to model the data distribution $p_{\theta}(\x)$ based on the Gaussian transitions.
In the geometric view of diffusion models~\cite{chung2022improving}, the transitions are typically described as iterative manifold transitions $\mathcal{M}_{t} \to \mathcal{M}_{t-1}, t=T,\cdots, 1$, moving from the noisy manifold $\mathcal{M}_T$ to the clean manifold $\mathcal{M}_0$.

Diffusion model-based inverse problem solvers (DIS)~\citep{kawar2022denoising, chung2023diffusion, song2023pseudoinverseguided, wang2023zeroshot, chung2024decomposed} aim to guide manifold transitions to sample from the posterior distribution $p_{\theta}(\x|\y)$, which represents sampling $\x$ from the measurement $\y$ obtained from the forward model $\Ac(\x)$.
In Bayesian inference, the posterior distribution, $p_\theta(\x|\y) \propto p_\theta(\x)p(\y|\x)$ is decomposed into the likelihood $p(\y|\x)$, representing the probability of observing $\y$ given $\x$, and the prior data distribution $p_\theta(\x)$.
This decomposition enables posterior sampling by combining diffusion sampling with iterative guidance using the forward model $\Ac$ and measurement $\y$.
This approach provides sophisticated, precise solutions to complex inverse problems, leveraging the power and flexibility of diffusion models in practical applications, such as deblurring, super-resolution, inpainting, colorization, compressed sensing, and so on.

\noindent\textbf{DIS using latent diffusion models (LDIS).}
Most DIS~\cite{kawar2022denoising, chung2023diffusion, song2023pseudoinverseguided, wang2023zeroshot, chung2024decomposed} use {\em pixel}-space diffusion models, which facilitate easy integration of the forward model $\Ac$ and measurement $\y$, as both are defined in pixel space.
Integrating forward models in latent space presents more challenges.
{\em Latent}-space methods~\cite{rout2024solving, kim2023regularization, chung2023prompt} calculate data consistency terms after decoding the denoised latent representation, then update these guidances within the latent space.

During this process, VAE mapping errors accumulate in iterative sampling, causing the representation to drift from the clean manifold $\mathcal{M}_0$.
Additionally, most latent diffusion models provide a text-conditioned prior distribution $p_\theta(\x|c_\text{text})$, which is challenging to implement in cases where text conditioning ($c_\text{text}$) is unavailable. 
As a result, {\em latent}-space methods prioritize two main goals: (i) managing text embeddings effectively and (ii) preserving the updated latent close to the clean manifold $\mathcal{M}_0$.

For text embeddings, PSLD~\cite{rout2024solving} uses only null-text, while TReg~\cite{kim2023regularization} and P2L~\cite{chung2023prompt} apply either null-text optimization or text optimization to enhance the reconstruction performance.
To maintain the updated latent representation quality, the regularization term for aligning pixel and latent spaces is used to enforce latent feasibility~\cite{rout2024solving, chung2023prompt, kim2023regularization}.

\noindent\textbf{Solving video inverse problems using DIS.}
Recently, several extensions~\cite{kwon2024solving, yeh2024diffir2vr, daras2024warped} from DIS have been introduced to address video inverse problems. 
A straightforward application of image diffusion models to video, processing frames individually, often disrupts temporal consistency. 
These approaches maintain temporal coherence by employing a batch-consistent sampling strategy~\cite{kwon2024solving} and leveraging optical flow guidance to warp either latent representations~\cite{yeh2024diffir2vr} or the noise prior~\cite{daras2024warped}.

While these innovative approaches allow powerful image generative models~\cite{dhariwal2021diffusion, rombach2022high, podell2023sdxl} to address video inverse problems with reduced computational demands, there is still room for improvement. 
A key limitation of optical flow-based methods~\cite{yeh2024diffir2vr, daras2024warped} is their heavy reliance on the accuracy of the optical flow estimation module~\cite{teed2020raft, xu2022gmflow}. 
This dependency becomes particularly problematic in scenarios involving severe degradations, which can hinder the estimation process and limit the methods' applicability to broader restoration tasks.
Moreover, these approaches often require task-specific restoration modules~\cite{yeh2024diffir2vr} or fine-tuning of the diffusion model~\cite{daras2024warped}. 

In contrast, the batch-consistent sampling strategy~\cite{kwon2024solving} has demonstrated effectiveness in addressing various spatio-temporal degradations without the need for task-specific training or model fine-tuning. 
However, this method utilizes the unconditional {\em pixel}-space diffusion model provided by ADM~\cite{dhariwal2021diffusion}, and its extension to latent diffusion models remains unexplored.

\begin{figure*}[ht]
  \centering
  % \vspace{-0.5cm}
    \centerline{{\includegraphics[width=\linewidth]{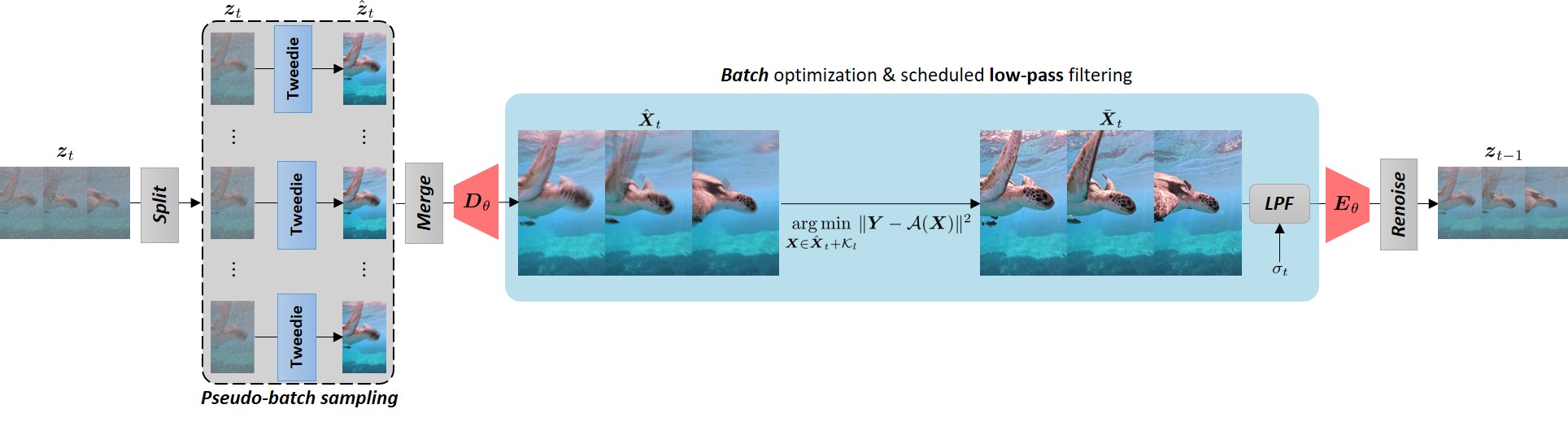}}}
    \vspace{-0.3cm}
    \caption{Illustration of VISION-XL sampling at timestep $t$: $\z_t$ is split into individual frames and denoised in parallel using Tweedie’s formula. The denoised latents $\hat{\z}_t$ are then merged and decoded. The decoded batch $\hat{\X}_t$ is optimized to enforce the data consistency, followed by low-pass filtered encoding and re-noising to obtain $\z_{t-1}$.}
    \label{fig:geometry2}
    \vspace{-0.5cm}
\end{figure*}

\section{High Definition Video Inverse Solver Using Latent Diffusion Models}

This section introduces a novel approach for reconstructing high-definition videos that include various spatio-temporal degradations. The overall pipeline of the algorithm is illustrated in Fig.~\ref{fig:geometry2}.

Consider the spatio-temporal degradation process is formulated as:
\begin{align}
\label{eq:forward}
\Y = \Ac(\X) = \Ac([\X[1], \cdots, \X[N]])
\end{align}
where $\Y$ denotes the measurement, $\X[n]$ denotes the $n$-th frame ground-truth frame, $N$ is the number of video frames, and $\Ac$ refers to the operator describing the spatio-temporal degradation process.

Our approach begins by inverting the measurement frames, denoted as $\Y$, to initialize the informative latents $\z_{\tau}$, enhancing batch-wise consistency (Step 1).
Next, we construct the corresponding denoised batch $\hat{\X}_{\tau}$ by sampling each latent in parallel using Tweedie's formula~\cite{efron2011tweedie}, followed by decoding (Step 2).
In Step 3, the corresponding denoised batch $\hat{\X}_{\tau}$ is further refined by applying $l$-step conjugate gradient (CG) optimization~\cite{chung2024decomposed, kwon2024solving} to enforce the data consistency from spatio-temporal degradation $\Ac$.
In Step 4, we apply a scheduled low-pass filter to the updated batch $\bar{\X}_{\tau}$ inspired by the frequency-based analysis of spectral diffusion~\cite{yang2023diffusion}. $\bar{\X}_{\tau}$ is then re-encoded into latent space to form $\bar{\z}_{\tau}$.
Finally, we obtain the one-step denoised latents $\z_{\tau-1}$ by adding noise to the encoded latents $\bar{\z}_{\tau}$ (Step 5). 
In the following, we provide a detailed description of each of these steps.

% \vspace*{0.25cm}

% \begin{figure}[t]
%   \centering
%     \centerline{{\includegraphics[width=0.9\linewidth]{figures/figure2.jpg}}}
%         \vspace{-0.3cm}
%     \caption{VISION-XL initialization using batch-consistent inversion of a measurement frame.}
%     \label{fig:geometry1}
%             \vspace{-0.3cm}
% \end{figure}

\noindent\textbf{Step 1: Initialize informative latents.}
One of our key insights is to initialize the informative latents by inverting the measurement frames, thereby enhancing batch-wise consistent initialization.
Although these latents cannot restore the ground-truth frames directly, inverted latent variables can inherit information from the measurement frame, providing good initializations~\cite{xiao2024dreamclean}.
Different from SVI~\cite{kwon2024solving}, which initializes sampling from a batch-wise synchronized uninformative Gaussian prior $\z_{T} \sim \mathcal{N}(0,I)$ as the initial sampling point, we replace $\z_{T}$ with the informative prior $\z_{\tau}$, defined as:
\begin{align}
%\label{eq:prior}
\z_0 = \boldsymbol{E}_{\theta}(\Y), \quad
\z_{\tau} = \text{DDIM}^{-1}(\z_0),
\end{align}
where $\boldsymbol{E}_{\theta}(\cdot)$ and $\text{DDIM}^{-1}(\cdot)$ denotes frame-wise encoding from pretrained VAE and DDIM inversion of timestep $\tau$, respectively.

\noindent\textbf{Step 2: Pseudo-batch sampling.}
After initialization, we guide the sampling path to ensure the data consistency condition. At timestep $0<t\leq\tau$, we sample denoise batch $\hat\z_{t}$ from given latents $\z_{t} := [\z_{t}[1] \cdots \z_{t}[N]]$ by using Tweedie's formula~\cite{efron2011tweedie}.
Unlike SVI~\cite{kwon2024solving}, we split latent frames to construct pseudo-batch and sample each frame in parallel, requiring memory for only a single frame during the sampling, as shown in Fig.~\ref{fig:geometry2}. Similarly, inversion is also conducted within the pseudo-batch framework.
This enables the recent advanced latent diffusion model to operate in this framework without a frame limit. 
As a proof-of-concept, we conduct experiments on 25-frame videos.

Specifically, consider a parallel sampling of latent diffusion models along the temporal direction:
\begin{align}\label{eq:batchdiff}
\boldsymbol{\mathcal{E}}_\theta^{(t)}(\z_t):= 
[\boldsymbol{\epsilon}_{\theta}^{(t)}(\z_t[1]) \cdots \boldsymbol{\epsilon}_{\theta}^{(t)}(\z_t[N])].
\end{align}
The denoised latents $\hat{\Z}_{t}$ are computed using Tweedie's formula~\cite{efron2011tweedie}:
\begin{align}
\label{eq:Tweedie}
    \hat{\z}_{t} = 
    \frac{1}{\sqrt{\bar\alpha_t}}\left(\z_t - \sqrt{1 - \bar\alpha_t} \boldsymbol{\mathcal{E}}_{\theta}^{(t)}(\z_t)  \right),
\end{align}
where $\bar\alpha_t$ is the noise schedule defined in the Gaussian process of diffusion models~\cite{ho2020denoising, nichol2021improved}.
Then denoised batch $\hat{\X}_{t}$ is decoded from the denoised latents $\hat{\z}_{t}$ using VAE decoder $\boldsymbol{D}_{\theta}$:
\begin{align}
\label{eq:decode}
    \hat{\X}_{t} = \boldsymbol{D}_{\theta}(\hat{\z}_{t}) := [\boldsymbol{D}_{\theta}(\hat{\z}_{t}[1]) \cdots \boldsymbol{D}_{\theta}(\hat{\z}_{t}[N])].
\end{align}

\noindent\textbf{Step 3: Batch optimization in \textit{pixel}-space.}
The denoised batch $\hat{\X}_{t}$ is then refined as a whole by applying the $l$-step CG optimization to enhance the data consistency from the measurement $\Y$ and spatio-temporal degradation $\Ac$.
This can be formally represented by
\begin{align}
\label{eq:CG}
   \bar\X_t :=  \argmin_{\X \in \hat\X_t + \boldsymbol{\mathcal K}_l}\|\Y - \Ac(\X)\|^2
\end{align}
where $\boldsymbol{\mathcal K}_l$ denotes the $l$-dimensional Kyrlov subspace associated with the given inverse problem \citep{chung2024decomposed}.
The multistep CG allows each temporal frame to be diversified, enhancing data consistency and achieving faster convergence without requiring memory-intensive gradient calculations~\cite{chung2023diffusion}.

\noindent\textbf{Step 4: Low-pass filtered encoding.}
Recent frequency-based analyses of diffusion models~\cite{karras2022elucidating, yang2023diffusion} suggest that optimal denoisers first recover low-frequency components in early denoising stages, while high-frequency details added progressively in later stages.
Building on these findings, we observed that applying a scheduled low-pass filter to the updated batch $\bar\X_t$ in early stages effectively removes undesired artifact caused by VAE error accumulation, resulting in more natural and refined outputs. 

Based on the observation that the denoiser restores high-frequency details as the noise scale $\sqrt{1-\bar\alpha_t}$ decreases, we set the filter width $\sigma_t$ to be proportional to the noise scale~\cite{nichol2021improved}, defined as $\sigma_t := \lambda \sqrt{1-\bar\alpha_t}$, which goes to zero at $t\rightarrow 0$.
After applying the low-pass filter $\boldsymbol{h}_{\sigma_t}$, we re-encode $\bar\X_t$ into the latent space.
Specifically, the re-encoded latents are given by:
\begin{gather}
\label{eq:lpf_encode}
    \bar\X_{t} \leftarrow \bar\X_{t} \ast \boldsymbol{h}_{\sigma_t}, \\
    \bar\z_t = \boldsymbol{E}_{\theta}(\bar\X_{t}) := [\boldsymbol{E}_{\theta}(\bar\X_{t}[1]) \cdots \boldsymbol{E}_{\theta}(\bar\X_{t}[N])],
\end{gather}
where $\boldsymbol{E}_{\theta}$ denotes the VAE encoder.

\noindent\textbf{Step 5: Renoising.}
After encoding, updated latents $\bar\z_t$ are renoised as:
\begin{align}
\label{eq:renoise}
   \z_{t-1} = \sqrt{\bar\alpha_{t-1}}\bar\z_{t} + \sqrt{1 - \bar\alpha_{t-1}} \boldsymbol{\mathcal{E}}_{t},
\end{align}
where $\boldsymbol{\mathcal{E}}_{t}$ is composed of batch-consistent stochastic noise~\cite{kwon2024solving} and deterministic noise~\cite{song2021denoising}.

\begin{figure}[!t]
  \centering
  \vspace{-0.3cm}
    \centerline{{\includegraphics[width=0.8\linewidth]{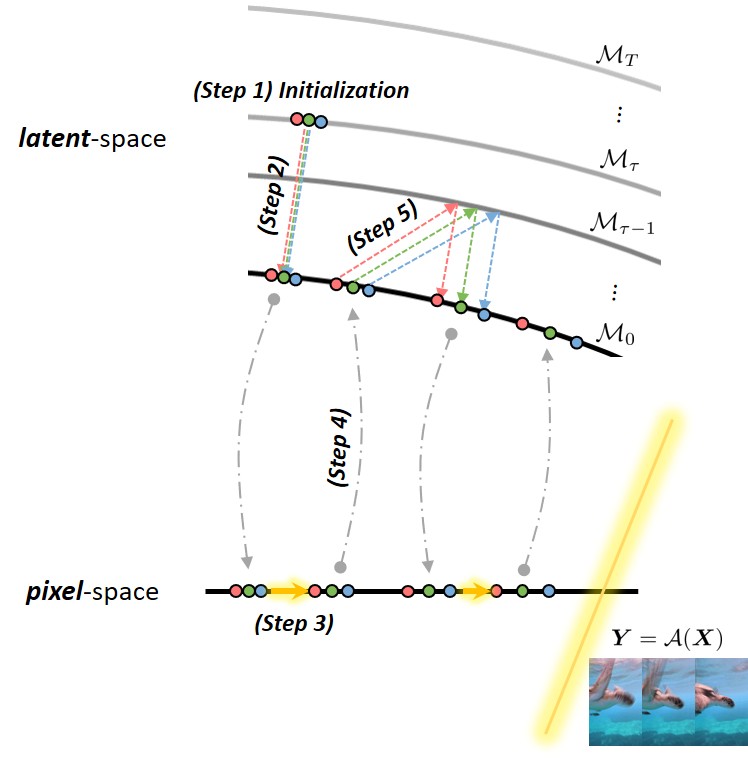}}}
    \vspace{-0.3cm}
    \caption{Geometric illustration of the sampling path evolution. (Step 1) Initialize latent $\z_{\tau}$. (Step 2) Project onto $\mathcal{M}_0$ via pseudo-batch sampling and decode to pixel space. (Step 3) Optimize for measurement consistency $\Y=\Ac(\X)$. (Step 4) Apply a scheduled low-pass filter and encode back to latent space. (Step 5) Renoise to $\mathcal{M}_{\tau-1}$.}
    \vspace{-0.1cm}
    \label{fig:geometry3}
\end{figure}

% In summary, the proposed method initializes informative latents and iteratively refines the decoded batch through multistep CG to meet spatio-temporal data consistency.
% We then apply low-pass filtered encoding to improve reconstruction quality.
% Extensive experimental results demonstrate that our method effectively addresses various spatio-temporal degradations, achieving state-of-the-art video reconstructions for high-definition videos and supporting a range of ratios previously unaddressed in other works.
In the geometric view of diffusion models~\cite{chung2022improving}, the sampling path evolves as illustrated in Fig.~\ref{fig:geometry3}.
The initialized latent $\z_{\tau}$ is projected onto the clean manifold $\mathcal{M}_0$ using Tweedie's formula. The projected latent is then decoded into the pixel space and refined through multi-step CG to satisfy the data consistency constraint $\Y=\Ac(\X)$.
Next, a scheduled low-pass filter is applied to reduce VAE error accumulation and keep the encoding close to $\mathcal{M}_0$. Finally, the encoded latents are re-noised to transition back to $\mathcal{M}_{\tau-1}$, and this process iterates until the final state converges to the clean manifold $\mathcal{M}_0$.
The complete algorithm is provided in Algorithm~\ref{alg:ours}. 

% \vspace{-0.2cm}

\begin{algorithm}
   \centering
   \caption{High-definition video inverse problem solver using latent diffusion models}\label{alg:ours}
   \begin{algorithmic}[1]
        \Require $\boldsymbol{\mathcal{E}}_{\theta}^{(t)}, \boldsymbol{E}_{\theta}, \boldsymbol{D}_{\theta}, \Y, \Ac, \tau, l, \sigma_t, \{\alpha_t\}^T_{t=1}$
        \State $\z_0 \leftarrow \boldsymbol{E}_{\theta}(\Y)$
        \State $\z_{\tau} \leftarrow \text{DDIM}^{-1}(\z_0)$ \Comment{\textbf{Step 1}}
        \For{$t = \tau:2$} \do \\
        \State $\hat{\z}_{t} \leftarrow \left(\z_t - \sqrt{1 - \bar\alpha_t} \boldsymbol{\mathcal{E}}_{\theta}^{(t)}(\z_t)  \right) / \sqrt{\bar\alpha_t}$ \Comment{\textbf{Step 2}}
        \State $\hat{\X}_{t} \leftarrow \boldsymbol{D}_{\theta}(\hat{\z}_{t})$
        \State $\bar\X_t :=  \argmin_{\X \in \hat\X_t + \boldsymbol{\mathcal K}_l}\|\Y - \Ac(\X)\|^2$ \Comment{\textbf{Step 3}}
        \State $\bar\X_{t} \leftarrow \bar\X_{t} \ast \boldsymbol{h}_{\sigma_t}$ \Comment{\textbf{Step 4}}
        \State $\bar\z_t = \boldsymbol{E}_{\theta}(\bar\X_{t})$
        \State $\z_{t-1} = \sqrt{\bar\alpha_{t-1}}\bar\z_{t} + \sqrt{1 - \bar\alpha_{t-1}} \boldsymbol{\mathcal{E}}_{t}$ \Comment{\textbf{Step 5}}
        \EndFor
        \State $\z_0 \leftarrow \left(\z_1 - \sqrt{1 - \bar\alpha_1} \boldsymbol{\mathcal{E}}_{\theta}^{(1)}(\z_1)  \right) / \sqrt{\bar\alpha_1}$
        \State \textbf{return} $\Z_0$
   \end{algorithmic}
\end{algorithm}

% \vspace{-0.7cm}

\section{Experimental Results}
\subsection{Experimental setup}
\noindent\textbf{Dataset.}
We used four high-resolution (with resolutions exceeding 1080p) video datasets for evaluation, sourced from the DAVIS dataset \cite{pont20172017} and the Pexels dataset\footnote{\url{https://www.pexels.com/}}.
A subset of 100 videos from the DAVIS dataset is resized to 768×1280 resolution and consists of 25 frames, originally provided in landscape orientation.
The Pexels dataset is a large, open-source collection of high-resolution stock videos and images, widely used for creative and research purposes. 
For the Pexels subset, we collect a total of 120 videos: 45 in landscape orientation (Pexels (landscape)), 45 in vertical orientation (Pexels (vertical)), and 30 in square orientation (Pexels (square)). 
These subsets are resized to resolutions of 768×1280 for landscape, 1280×768 for vertical, and 1024×1024 for square orientations, with each video consisting of 25 frames.
%We have open-sourced all datasets for easy reproducibility.

\noindent\textbf{Inverse problems.}
We test our method on the following spatial degradations: \textbf{1) Deblur}: Gaussian deblurring from an
image convolved with a 61×61 size Gaussian kernel with $\sigma$=3.0, \textbf{2) SR}: Super-resolution from $\times$4 average pooling, \textbf{3) Inpaint}: Inpainting from 50\% random masking. Furthermore, test our method on the following spatio-temporal degradations: \textbf{4) Deblur+}: Deblur + 7-frame averaging using temporal uniform blur kernel as used in \cite{kwon2024solving}, \textbf{5) SR+}: SR + 7-frame averaging, and \textbf{6) Inpaint+}: Inpaint + 7-frame averaging.

\begin{table*}[t]
    \centering
    \vspace{-0.3cm}
    \resizebox{\linewidth}{!}{
    \begin{tabular}{cccccccccccccccccc}
        \toprule
        & & 
        \multicolumn{4}{c}{\textbf{DAVIS}} & \multicolumn{4}{c}{\textbf{Pexels (landscape)}} & 
        \multicolumn{4}{c}{\textbf{Pexels (vertical)}} & \multicolumn{4}{c}{\textbf{Pexels (square)}} \\
        \cmidrule(lr){3-6} \cmidrule(lr){7-10} \cmidrule(lr){11-14} \cmidrule(lr){15-18}
        Task & Method & FVD$\downarrow$ & LPIPS$\downarrow$ & PSNR$\uparrow$ & SSIM$\uparrow$ & FVD$\downarrow$ & LPIPS$\downarrow$ & PSNR$\uparrow$ & SSIM$\uparrow$ & FVD$\downarrow$ & LPIPS$\downarrow$ & PSNR$\uparrow$ & SSIM$\uparrow$ & FVD$\downarrow$ & LPIPS$\downarrow$ & PSNR$\uparrow$ & SSIM$\uparrow$ \\
        \midrule

        \multirow{4}{*}{\textbf{Deblur+}}
        & ADMM-TV & 1512 & 0.397 & 24.30 & 0.742 & 671.8 & 0.284 & 29.42 & 0.818 & 709.2 & 0.237 & 30.37 & 0.847 & 569.5 & 0.224 & 30.68 & 0.856 \\
        & SVI~\cite{kwon2024solving} & \underline{638.9} & \underline{0.322} & \underline{28.04} & \underline{0.799} & \underline{830.5} & \underline{0.265} & \underline{30.42} & \underline{0.831} & \underline{656.2} & \underline{0.239} & \underline{30.40} & \underline{0.856} & \underline{499.4} & \underline{0.221} & \underline{31.59} & \underline{0.862} \\
        & DiffIR2VR~\cite{yeh2024diffir2vr} & - & - & - & - &  - & - & - & - &  - & - & - & - &  - & - & - & - \\
        \rowcolor{lightgray}
        \cellcolor{white} & Ours & \textbf{228.6} & \textbf{0.292} & \textbf{28.76} & \textbf{0.807} & \textbf{196.4} & \textbf{0.249} & \textbf{31.11} & \textbf{0.839} & \textbf{209.9} & \textbf{0.224} & \textbf{31.87} & \textbf{0.860} & \textbf{157.9} & \textbf{0.217} & \textbf{32.40} & \textbf{0.864} \\
        \midrule

        \multirow{4}{*}{\textbf{SR+}} 
        & ADMM-TV & 1429 & 0.359 & 24.23 & 0.740 & 634.1 & 0.279 & 29.01 & 0.820 & 669.2 & 0.322 & 29.71 & 0.836 & 545.9 & 0.306 & 30.08 & 0.838 \\
        & SVI~\cite{kwon2024solving} & \underline{223.4} & \underline{0.234} & \underline{29.00} & \underline{0.812} & \underline{386.8} & \underline{0.265} & \underline{30.70} & \underline{0.831} & \underline{558.9} & \underline{0.261} & \underline{30.48} & \underline{0.842} & \underline{313.2} & \underline{0.265} & \underline{31.18} & \underline{0.847} \\
        & DiffIR2VR~\cite{yeh2024diffir2vr} & - & - & - & - &  - & - & - & - &  - & - & - & - &  - & - & - & - \\
        \rowcolor{lightgray}
        \cellcolor{white} & Ours & \textbf{158.5} & \textbf{0.244} & \textbf{29.18} & \textbf{0.818} & \textbf{166.2} & \textbf{0.246} & \textbf{30.82} & \textbf{0.832} & \textbf{173.1} & \textbf{0.229} & \textbf{31.38} & \textbf{0.847} & \textbf{138.8} & \textbf{0.220} & \textbf{31.90} & \textbf{0.856} \\
        \midrule

        \multirow{4}{*}{\textbf{Inpaint+}}
        & ADMM-TV & 1848 & 0.339 & 24.16 & 0.762 & 797.8 & 0.292 & 29.15 & 0.778 & 805.4 & 0.258 & 29.36 & 0.805 & 652.4 & 0.268 & 30.39 & 0.804 \\
        & SVI~\cite{kwon2024solving} & \textbf{208.6} & \textbf{0.238} & \textbf{29.60} & \textbf{0.848} & \underline{269.3} & \underline{0.250} & \underline{29.92} & \underline{0.826} & \underline{428.1} & \underline{0.246} & \underline{30.27} & \underline{0.838} & \underline{206.9} & \underline{0.238} & \underline{31.12} & \underline{0.847} \\
        & DiffIR2VR~\cite{yeh2024diffir2vr} & - & - & - & - &  - & - & - & - &  - & - & - & - &  - & - & - & - \\
        \rowcolor{lightgray}
        \cellcolor{white} & Ours & \underline{241.1} & \underline{0.242} & \underline{28.81} & \underline{0.815} & \textbf{222.4} & \textbf{0.216} & \textbf{30.13} & \textbf{0.828} & \textbf{240.4} & \textbf{0.201} & \textbf{30.98} & \textbf{0.845} & \textbf{164.9} & \textbf{0.230} & \textbf{31.44} & \textbf{0.853} \\
        \midrule
        
        \multirow{4}{*}{\textbf{Deblur}} 
        & ADMM-TV & 169.0 & 0.232 & 30.49 & 0.873 & 192.2 & 0.263 & 31.67 & 0.842 & 185.7 & 0.181 & 32.19 & 0.861 & 199.6 & 0.257 & 30.98 & 0.875 \\
        & SVI~\cite{kwon2024solving} & \underline{99.50} & \underline{0.176} & \underline{31.70} & \underline{0.875} & \underline{153.8} & \underline{0.212} & \underline{32.44} & \underline{0.862} & \underline{160.2} & \underline{0.174} & \underline{33.35} & \underline{0.866} & \underline{116.6} & \underline{0.207} & \underline{33.51} & \underline{0.885} \\
        & DiffIR2VR~\cite{yeh2024diffir2vr} & - & - & - & - &  - & - & - & - &  - & - & - & - &  - & - & - & - \\
        \rowcolor{lightgray}
        \cellcolor{white} & Ours & \textbf{72.03} & \textbf{0.171} & \textbf{31.94} & \textbf{0.880} & \textbf{96.16} & \textbf{0.184} & \textbf{32.83} & \textbf{0.876} & \textbf{89.39} & \textbf{0.170} & \textbf{33.55} & \textbf{0.888} & \textbf{91.55} & \textbf{0.157} & \textbf{34.10} & \textbf{0.896} \\
        \midrule

        \multirow{4}{*}{\textbf{SR}} 
        & ADMM-TV & 285.6 & 0.221 & 26.65 & 0.788 & 301.5 & 0.222 & 27.91 & 0.769 & 298.7 & 0.213 & 27.12 & 0.786 & 209.4 & 0.257 & 28.74 & 0.798 \\
        & SVI~\cite{kwon2024solving} & \underline{176.1} & \underline{0.176} & \underline{29.28} & \underline{0.815} & \underline{201.6} & \underline{0.202} & \underline{31.00} & \underline{0.836} & \underline{219.3} & \underline{0.200} & \underline{30.59} & \underline{0.847} & \underline{167.9} & \underline{0.206} & \underline{31.83} & \underline{0.851} \\
        & DiffIR2VR~\cite{yeh2024diffir2vr} & 319.6 & 0.238 & 26.29 & 0.738 & 412.6 & 0.244 & 27.18 & 0.742 & 511.6 & 0.214 & 26.91 & 0.752 & 412.0 & 0.296 & 27.76 & 0.741 \\
        \rowcolor{lightgray}
        \cellcolor{white} & Ours & \textbf{81.10} & \textbf{0.190} & \textbf{30.27} & \textbf{0.848} & \textbf{104.3} & \textbf{0.185} & \textbf{31.84} & \textbf{0.858} & \textbf{98.57} & \textbf{0.158} & \textbf{32.54} & \textbf{0.876} & \textbf{98.41} & \textbf{0.138} & \textbf{33.02} & \textbf{0.886} \\
        \midrule

        \multirow{4}{*}{\textbf{Inpaint}} 
        & ADMM-TV & 212.0 & 0.315 & 28.42 & 0.797 & 270.9 & 0.316 & 29.14 & 0.794 & 270.3 & 0.212 & 29.84 & 0.803 & 264.7 & 0.310 & 30.46 & 0.809 \\
        & SVI~\cite{kwon2024solving} & \textbf{143.5} & \textbf{0.177} & \textbf{30.20} & \textbf{0.858} & \underline{159.4} & \underline{0.233} & \underline{30.09} & \underline{0.827} & \textbf{164.3} & \textbf{0.132} & \textbf{31.19} & \textbf{0.847} & \underline{139.3} & \underline{0.225} & \underline{31.57} & \underline{0.855} \\
        & DiffIR2VR~\cite{yeh2024diffir2vr} & - & - & - & - &  - & - & - & - &  - & - & - & - &  - & - & - & - \\
        \rowcolor{lightgray}
        \cellcolor{white} & Ours & \underline{143.6} & \underline{0.209} & \underline{29.74} & \underline{0.835} & \textbf{158.7} & \textbf{0.208} & \textbf{30.41} & \textbf{0.834} & \underline{174.0} & \underline{0.195} & \underline{31.15} & \textbf{0.847} & \textbf{125.4} & \textbf{0.216} & \textbf{31.63} & \textbf{0.859} \\

        \bottomrule
    \end{tabular}
    }
    \vspace{-0.2cm}
    \caption{Quantitative evaluation (FVD, LPIPS, PSNR, SSIM) of solving spatio-temporal inverse problems across DAVIS, Pexels dataset with multiple aspect ratios (landscape, vertical, square). {\textbf{Bold} denotes the best results and \underline{underline} indicates the runner-up. Notably, DiffIR2VR~\cite{yeh2024diffir2vr} is only capable of restoring SR among our experimental tasks, highlighting the broader task generalizability of our approach.}}
    \label{table:main}
\end{table*}

\begin{figure*}[!ht]
  \centering
  \vspace{-0.2cm}
    \centerline{{\includegraphics[width=\linewidth]{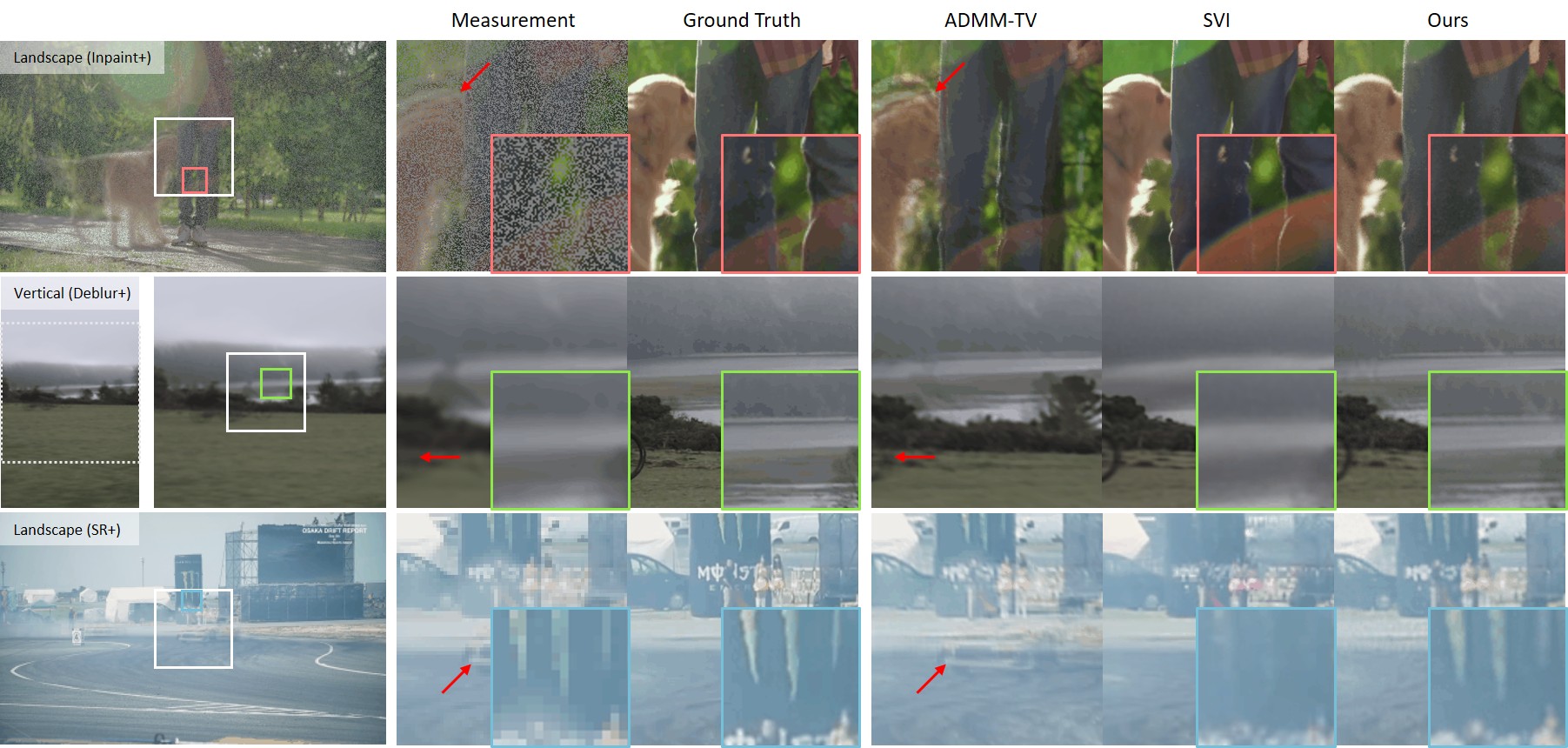}}}
    \vspace{-0.2cm}
    \caption{Qualitative evaluation of solving spatio-temporal inverse problems across DAVIS, Pexels dataset with multiple aspect ratios. {Notably, ADMM-TV fails to remove ghosting artifacts caused by temporal degradation (red arrows), while SVI produces excessive intensity fluctuations (red box) or blurred information restoration (green and blue boxes).}}
    \vspace{-0.3cm}
    \label{fig:result1}
\end{figure*}

\begin{figure*}[!ht]
  \centering
   \vspace{-0.3cm}
    \centerline{{\includegraphics[width=\linewidth]{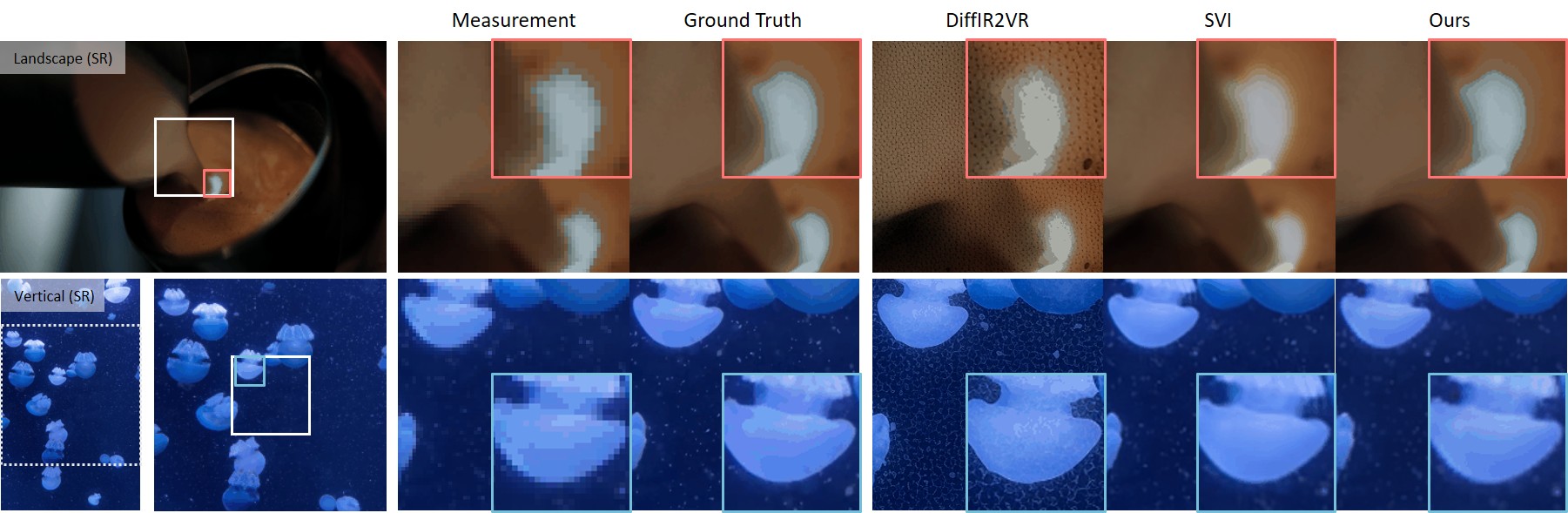}}}
     \vspace{-0.2cm}
    \caption{Qualitative evaluation of SR ($\times$4) performance across multiple aspect ratios (landscape, vertical). {DiffIR2VR often produces unwanted artifacts in the background (red and blue boxes), while SVI inaccurately restores intensity (red box), leading to frame-wise fluctuations.}}
    \label{fig:result2}
    \vspace{-0.5cm}
\end{figure*}

\begin{table}[ht]
\centering
    \resizebox{0.6\linewidth}{!}{
        \begin{tabular}{ccc}
        \toprule
        Method & Time [min] & Memory (GB) \\
        \cmidrule(lr){1-1}\cmidrule(lr){2-3}
        SVI~\cite{kwon2024solving} & 15 & 18.5 \\
        DiffIR2VR~\cite{yeh2024diffir2vr} & \underline{4.7} & \underline{13.6} \\
        Ours & \textbf{2.5} & \textbf{12.7} \\
        \bottomrule
        \end{tabular}
    }
    \vspace{-0.2cm}
\caption{Comparison of total sampling time and memory efficiency for solving video inverse problems on a single 25-frame video at 768×1280 resolution. {\textbf{Bold} denotes the best results and \underline{underline} indicates the runner-up.}}
\label{table:time}
\vspace{-0.5cm}
\end{table}

\noindent\textbf{Baseline comparison.}
The primary objective of this study is to improve the performance of video inverse problem solvers through latent image diffusion models. Thus, our evaluation primarily compares video inverse problem solvers using image diffusion models.
As a recently emerging field, only a few methods are available: SVI~\cite{kwon2024solving}, DiffIR2VR~\cite{yeh2024diffir2vr}, and Warped Diffusion~\cite{daras2024warped}.
Notably, DiffIR2VR and Warped Diffusion cannot address spatio-temporal degradations, and DiffIR2VR only supports SR among the inverse problems we address in this paper.
We conducted comparisons with SVI and DiffIR2VR but excluded Warped Diffusion, as it is not currently open-source. 
SVI officially supports a resolution of 256×256, while DiffIR2VR supports 480×854. 
To ensure fair comparisons with identical resolutions, we used patch reconstruction. 
Additionally, we included a comparison with the classical optimization method ADMM-TV, following the protocol established by SVI~\cite{kwon2024solving}.

For quantitative comparison, we focus on two widely used standard metrics: peak signal-to-noise ratio (PSNR) and structural similarity index (SSIM)~\cite{wang2004image}. 
Additionally, we evaluate two perceptual metrics: Learned Perceptual Image Patch Similarity (LPIPS)~\cite{zhang2018perceptual} and Fréchet Video Distance (FVD)~\cite{unterthiner2019fvd}. 
For computing the metrics, we follow the protocol from the open-source project\footnote{\url{https://github.com/JunyaoHu/common_metrics_on_video_quality}}.

\noindent\textbf{Implementation details.}
While our method is applicable to general latent diffusion models, we use Stable Diffusion XL 1.0 (SDXL)~\cite{podell2023sdxl}—the current state-of-the-art text-to-image diffusion model—as a proof of concept in this paper.
For all experiments, we employ $T=25$, $\tau=0.3 T$, $\lambda=2$, and $l=10$. These optimal values are obtained through extensive ablation studies and the results are described in Sec.~\ref{sec:ablation}.
To reduce undesired guidance from the text condition $\boldsymbol{c}_{\text{text}}$, we use a null-text condition, $\boldsymbol{c}_{\varnothing}$.
All experiments were done on a single NVIDIA 4090 GPU.

\subsection{Results}
Table~\ref{table:main} presents a quantitative comparison across various spatio-temporal inverse problems. The proposed method consistently outperforms baseline approaches in most metrics, particularly in addressing spatio-temporal degradations. Notably, it achieves a significant reduction in FVD, indicating superior perceptual video quality compared to the runner-up across all datasets. This improvement is also evident in the qualitative results shown in Fig.~\ref{fig:result1}.
While SVI~\cite{kwon2024solving} effectively handles spatio-temporal degradations, it often struggles with temporal consistency, leading to issues such as inaccurate intensity restoration (first row) and loss of details (second and third rows). This suggests that batch-consistent noise initialization~\cite{kwon2024solving}, further examined in our ablation study (Table~\ref{table:ablation_1}), may be insufficient to fully preserve temporal consistency. Additionally, its patch reconstruction for high-resolution videos can introduce patch-wise inconsistencies, degrading overall performance.
The classical optimization method, ADMM-TV, effectively reconstructs static backgrounds and stationary objects but fails to remove ghosting artifacts caused by temporal degradations, as shown in Fig.~\ref{fig:result1}. This limitation is reflected in its lower performance metrics.
For the SR task (Fig.\ref{fig:result2}), DiffIR2VR\cite{yeh2024diffir2vr} often introduces unwanted artifacts in backgrounds or over-generates object details, likely due to inaccuracies in optical flow estimation.
{Notably, in pixel-wise random inpainting, latent diffusion methods may lose fine pixel-level details due to their encoded representations. However, leveraging the strong SDXL prior, our method achieves competitive inpainting performance with SVI, which employs a pixel-space diffusion model.}

While baseline methods encounter various challenges across different inverse problems, our approach demonstrates stable, high-quality reconstructions, as evidenced by the overall results. Further comparisons of total sampling time and memory consumption are shown in Table~\ref{table:time}, where our method achieves the highest efficiency in both sampling time and memory usage.

Additional visualizations, including reconstruction results for deblurring, inpainting, and other tasks, are available in video format for further evaluation:
{\small \textsf{\url{https://vision-xl.github.io/supple/}}}.

\subsection{Ablation studies} \label{sec:ablation}
{In this section, we analyze the key components of our method. To highlight their impact, we conduct an ablation study on the SR+ task in the Pexels (landscape) dataset, which involves significant spatio-temporal degradation.}
%, demonstrating that each proposed element is crucial for achieving state-of-the-art reconstruction of degraded high-definition videos. 
%In the following, we provide detailed ablation studies of key steps.

\begin{table}[ht]
\centering
    \resizebox{\linewidth}{!}{
        \begin{tabular}{cccccc}
        \toprule
        Initialization & Time [min] & FVD$\downarrow$ & LPIPS$\downarrow$ & PSNR$\uparrow$ & SSIM$\uparrow$ \\
        \cmidrule(lr){1-1}\cmidrule(lr){2-6}
        Random noise  & 8.3 & 1047 & 0.251  & 29.43  & 0.822  \\
        Batch-consistent noise~\cite{kwon2024solving} & 8.3 & 707.7 & 0.248 & 30.10 & 0.824 \\
        \cmidrule(lr){1-6}
        Pseudo-batch inversion ($\tau$: 0.15$T$)  & \textbf{1.3} & \underline{229.5} & 0.244 & 30.00 & 0.806 \\
        {\em Pseudo-batch inversion ($\tau$: 0.30$T$)}  & \underline{2.5} & \textbf{184.8} & \textbf{0.236} & \textbf{30.74}  & \underline{0.826} \\
        Pseudo-batch inversion ($\tau$: 0.45$T$)  & 3.8 & 288.7 & \underline{0.241} & \underline{30.70}  & \textbf{0.827} \\ 
        \bottomrule
        \end{tabular}
    }
\caption{Ablation study on the effect of the initializations.}
\label{table:ablation_1}
\end{table}

\begin{figure}[!t]
  \centering
    \centerline{{\includegraphics[width=\linewidth]{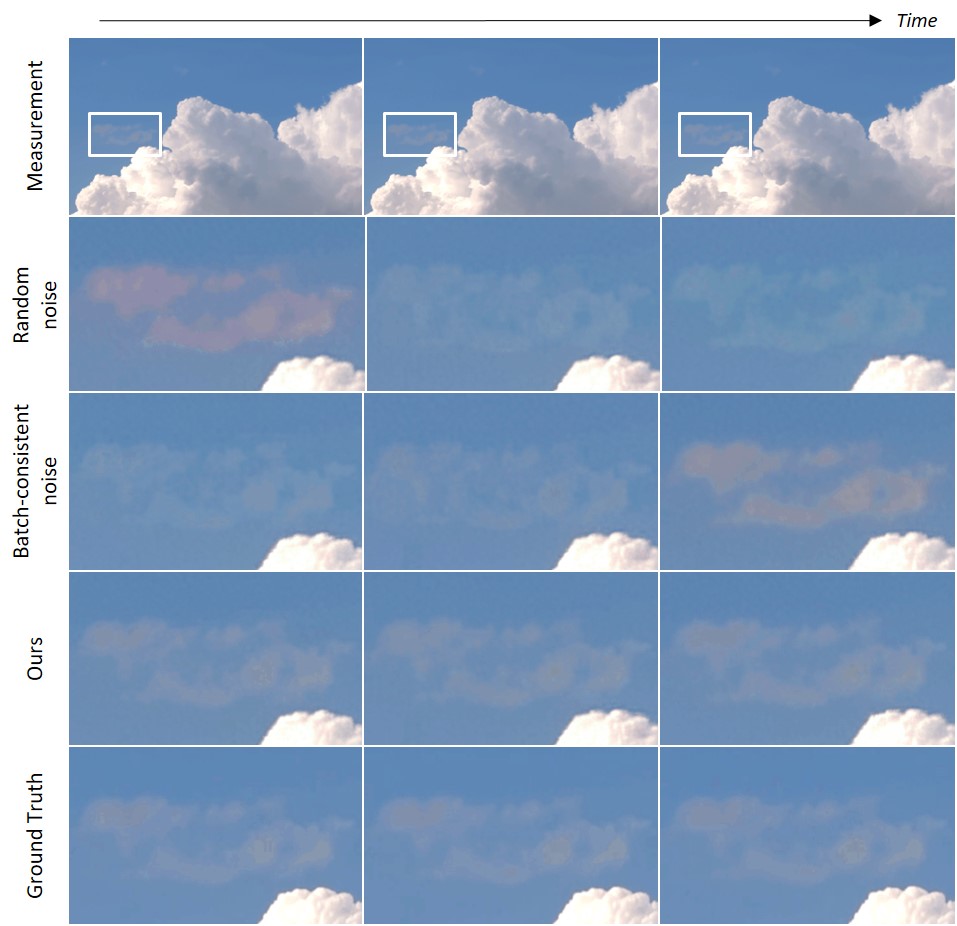}}}
    \caption{Ablation study on the effect of the initializations.}
    \label{fig:ablation1}
    % \vspace{-0.2cm}
\end{figure}

\noindent\textbf{Effect of initialization (in Step 1).}
In Table~\ref{table:ablation_1}, we conduct an ablation study to see the effect of pseudo-batch inversion for initialization.
From the table, we confirm that pseudo-batch inversion effectively extracts informative latents to reconstruct video evidenced by about 0.6dB and 1.3dB PSNR increase compared with batch-synchronized noise initialization used in SVI~\cite{kwon2024solving} and random noise initialization, respectively.
Notably, pseudo-batch inversion achieves $\times$3 lower FVD compared to the batch-synchronized noise initialization which indicates a significant improvement in temporal consistency.
It is also evident in the visualization of the ablation study shown in Fig.~\ref{fig:ablation1}.
The reconstruction results from random noise and batch-synchronized noise initialization fail to reconstruct the color of the cloud and are temporally inconsistent.
In contrast, our method successfully reconstructs the color of the cloud and temporally consistent results.
Furthermore, as shown in Table~\ref{table:ablation_1}, pseudo-batch inversion significantly improves sampling time efficiency. This is because it reduces the total sampling steps, and inversion does not require the measurement update process, which involves encoding and decoding. As a result, our method reconstructs high-definition video in under 6 seconds per frame on a single NVIDIA 4090 GPU.

\begin{table}[ht]
\centering
    \resizebox{0.65\linewidth}{!}{
        \begin{tabular}{ccccc}
        \toprule
        Update step $l$ & FVD$\downarrow$ & LPIPS$\downarrow$ & PSNR$\uparrow$ & SSIM$\uparrow$ \\
        \cmidrule(lr){1-1}\cmidrule(lr){2-5}
        1 & 1150 & 0.281  & 27.55  & 0.799  \\
        5 & \underline{241.0} & \textbf{0.197} & \underline{30.69} & \textbf{0.839} \\
        {\em 10} & \textbf{184.8} & \underline{0.236} & \textbf{30.74}  & \underline{0.826} \\
        20 & 486.1 & 0.470 & 28.16  & 0.690 \\
        \bottomrule
        \end{tabular}
    }
\caption{Ablation study on the effect of $l$.}
\label{table:ablation_2}
\vspace{-0.5cm}
\end{table}

\noindent\textbf{Effect of optimization step $l$ (in Step 3).}
In Table~\ref{table:ablation_2}, we present an ablation study on the effect of the CG update step $l$. 
The table confirms that the CG update is essential for enhancing data consistency.
We found that at least 5 iterations of CG updates yield satisfactory results, and 10 iterations produce the best results.

\begin{figure}[!t]
  \centering
    \centerline{{\includegraphics[width=\linewidth]{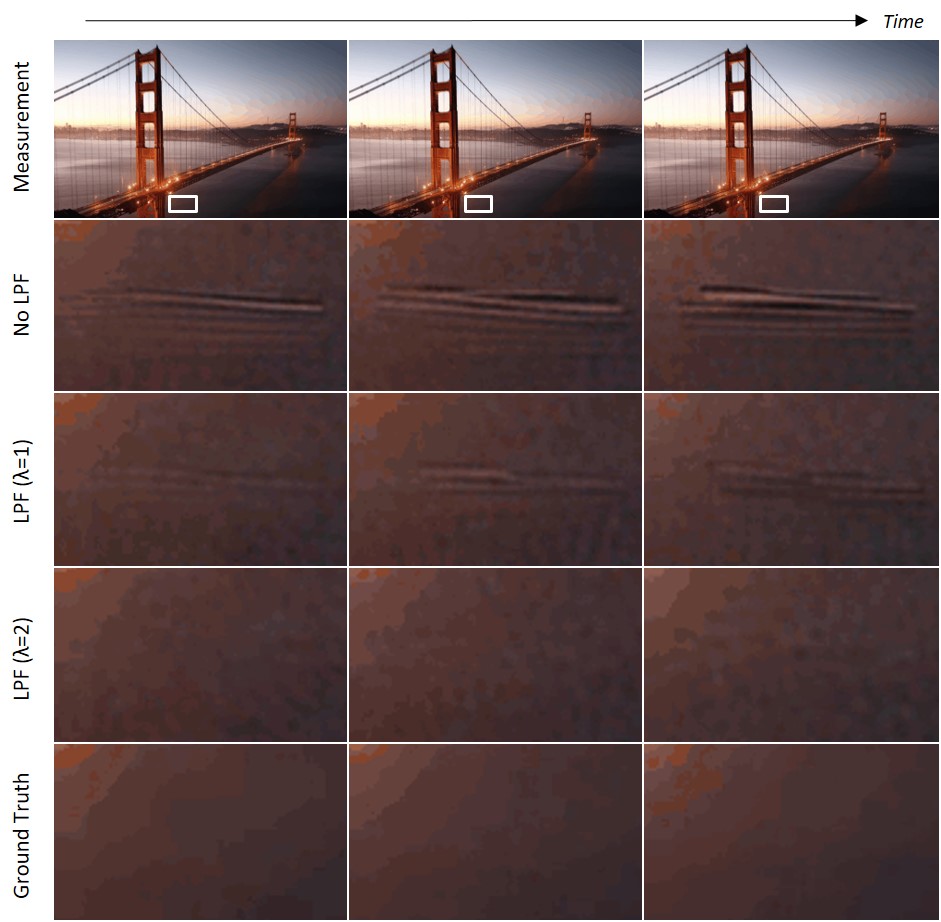}}}
    \caption{Ablation study on the effect of low-pass filtering.}
    \label{fig:ablation2}
    \vspace{-0.3cm}
\end{figure}

\begin{table}[ht]
\centering
    \resizebox{0.6\linewidth}{!}{
        \begin{tabular}{ccccc}
        \toprule
        LPF $\lambda$ & FVD$\downarrow$ & LPIPS$\downarrow$ & PSNR$\uparrow$ & SSIM$\uparrow$ \\
        \cmidrule(lr){1-1}\cmidrule(lr){2-5}
        No LPF & 209.7 & 0.273 & 29.92 & 0.797 \\
        \cmidrule(lr){1-5}
        1 & 186.3 & \underline{0.239} & 30.59 & 0.819 \\
        $\sqrt{2}$ & \underline{184.8} & \textbf{0.236} & 30.74 & 0.826 \\
        {\em 2} & \textbf{179.3} & 0.245 & \underline{30.81}  & \underline{0.832} \\
        2$\sqrt{2}$ & 191.4 & 0.262 & \textbf{30.89} & \textbf{0.837} \\
        \bottomrule
        \end{tabular}
    }
\caption{Ablation study on the effect of LPF $\lambda$.}
\label{table:ablation_3}
\vspace{-0.3cm}
\end{table}

\noindent\textbf{Effect of LPF $\lambda$ (in Step 4).}
Table~\ref{table:ablation_3} presents an ablation study on the effect of low-pass filtering. 
The results confirm that low-pass filtering enhances the reconstruction quality as evidenced by all metrics.
Specifically, low-pass filtering results in approximately a 30-point decrease in FVD and a 1.0dB increase in PSNR compared to the absence of low-pass filtering.
This improvement is also evident in the visualizations in Fig.~\ref{fig:ablation2}. 
In the second row of the figure, undesired artifacts appear when low-pass filtering is not applied. 
In contrast, as shown in the third and fourth rows, these artifacts are effectively removed as the parameter $\lambda$ increases.
From a frequency-based perspective, we believe that low-pass filtering effectively guides the updated latents to remain within the desired denoised manifold, $\mathcal{M}_0$, and helps to mitigate error accumulation from the VAE.

\section{Conclusion}
In this paper, we proposed a novel framework for addressing high-definition video inverse problems using latent diffusion models that introduce two new strategies.
First, a pseudo-batch consistent sampling strategy to manage intensive batch memory consumption with advanced latent diffusion models (e.g., SDXL).
To acquire a denoised batch, we conduct parallel sampling of each latents rather than batch sampling to efficiently manage the high memory consumption of advanced latent diffusion models.
Second, a pseudo-batch inversion for leveraging informative latent as initialization is proposed.
We confirmed that pseudo-batch inversion significantly improves reconstruction performance in both traditional and perceptual quality metrics.
Leveraging the powerful SDXL, our method achieves state-of-the-art performance across diverse spatio-temporal inverse problems, including challenging tasks such as the combination of frame averaging with deblurring, super-resolution, and inpainting. 
Importantly, our method supports multiple aspect ratios (landscape, vertical, and square), making it versatile for different video formats and delivering HD reconstructions {in under 6 seconds per frame} on a single NVIDIA 4090 GPU.
Overall, our framework not only enhances video reconstruction quality but also sets new standards for efficiency and flexibility in solving high-definition video inverse problems.

\newpage

{
    \small
    \bibliographystyle{ieeenat_fullname}
    \bibliography{main}
}

\clearpage
\setcounter{page}{1}
\maketitlesupplementary

\section{Experimental details}
\label{sec:appendix_A}
\subsection{Implementation of Comparative Methods}

\noindent\textbf{SVI~\cite{kwon2024solving}.}
For SVI, we use the official implementation\footnote{\url{https://github.com/solving-video-inverse/codes}}. Specifically, we utilize the same pre-trained image diffusion model, the unconditional ADM~\cite{dhariwal2021diffusion}.
Following the protocol described in \cite{kwon2024solving}, we set the parameters as $l = 5$ and $\eta = 0.8$ with 100 NFE sampling.
Since SVI officially supports a resolution of 256$\times$256, we applied patch-based reconstruction to ensure fair comparisons at identical resolutions.

\noindent\textbf{DiffIR2VR~\cite{yeh2024diffir2vr}.}
For DiffIR2VR, we use the official implementation\footnote{\url{https://github.com/jimmycv07/DiffIR2VR-Zero}}. Specifically, we employ the same pre-trained image diffusion model, Stable Diffusion 2.1~\cite{rombach2022high}.
DiffIR2VR is designed to support only super-resolution (SR) within the scope of our inverse problem. Therefore, we conducted SR experiments exclusively.
Following the protocol in \cite{yeh2024diffir2vr}, we set the upscale factor to 4 and the CFG scale factor to 4, with 50 NFE sampling.
DiffIR2VR officially supports resolutions of 480$\times$854. To ensure fair comparisons across resolutions, we applied patch-based reconstruction. For different aspect ratios, we set the resolution to 480$\times$854 for landscape orientation, 854$\times$480 for vertical orientation, and 512$\times$512 for square.

\noindent\textbf{ADMM-TV.}
Following the protocol in \cite{kwon2024solving}, we optimize the following objective:
\begin{align} \X^{\ast} = \underset{\X}{\mathrm{argmin}} \frac{1}{2} \| \Ac\X - \Y \|^2_2 + \lambda \| \boldsymbol{D} \X \|_{1}, \end{align}
where $\boldsymbol{D} = [\boldsymbol{D}_t, \boldsymbol{D}_h, \boldsymbol{D}_w]$ corresponds to the classical Total Variation (TV) regularization. Here, $t$, $h$, and $w$ represent temporal, height, and width directions, respectively.
The outer iterations of ADMM were set to 30, and the inner iterations of conjugate gradient (CG) were set to 20, consistent with the settings in \cite{kwon2024solving}. The parameters were set to $(\rho, \lambda) = (1, 0.001)$. The initial value of $\X$ was set to zero.

\section{Extension to blind video inverse problems}

Our method can be extended to address blind video inverse problems, such as blind video deblurring, demonstrated using the widely-used GoPro dataset~\citep{nah2017deep}. Here, we provide an example application of our method to blind video deblurring, showing its potential as a general framework for solving blind video inverse problems.

\begin{algorithm}[t]
   \centering
   \caption{Ours (blind) - Blind video deconvolution}\label{alg:ours_blind}
   \begin{algorithmic}[1]
        \Require $\boldsymbol{\mathcal{E}}_{\theta}^{(t)}, \boldsymbol{E}_{\theta}, \boldsymbol{D}_{\theta}, \Y, \tau, l, \sigma_t, \{\alpha_t\}^T_{t=1}, \addorange{f_{\phi}}$
        \State \addorange{$\X_{\text{pre}} \leftarrow f_{\phi}(\Y)$} \Comment{\addorange{Round 1} with estimated PSF}
        \State \addorange{$\boldsymbol{h}_{\sigma} \leftarrow \argmin_{\boldsymbol{h}_{\sigma}}\|\Y -  \X_{\text{pre}}\ast \boldsymbol{h}_{\sigma}\|^2$}
        \State $\z_{0} \leftarrow \boldsymbol{E}_{\theta}(\Y)$
        \State $\z_{\tau} \leftarrow \text{DDIM}^{-1}(\z_{0})$
        \For{$t = \tau:2$} \do \\
        \State $\hat{\z}_{t} \leftarrow \left(\z_t - \sqrt{1 - \bar\alpha_t} \boldsymbol{\mathcal{E}}_{\theta}^{(t)}(\z_t)  \right) / \sqrt{\bar\alpha_t}$
        \State $\hat{\X}_{t} \leftarrow \boldsymbol{D}_{\theta}(\hat{\z}_{t})$
        \State $\bar\X_t :=  \argmin_{\X \in \hat\X_t + \boldsymbol{\mathcal K}_l}\|\Y - \X \addorange{\ast \boldsymbol{h}_{\sigma}}\|^2$
        \State $\bar\X_{t} \leftarrow \bar\X_{t} \ast \boldsymbol{h}_{\sigma_t}$
        \State $\bar\z_t = \boldsymbol{E}_{\theta}(\bar\X_{t})$
        \State $\z_{t-1} = \sqrt{\bar\alpha_{t-1}}\bar\z_{t} + \sqrt{1 - \bar\alpha_{t-1}} \boldsymbol{\mathcal{E}}_{t}$ 
        \EndFor
        \State $\z_0 \leftarrow \left(\z_1 - \sqrt{1 - \bar\alpha_1} \boldsymbol{\mathcal{E}}_{\theta}^{(1)}(\z_1)  \right) / \sqrt{\bar\alpha_1}$
        \State \addblue{$\boldsymbol{h}_{\sigma} \leftarrow \argmin_{\boldsymbol{h}_{\sigma}}\|\Y -  \boldsymbol{D}_{\theta}(\z_0)\ast \boldsymbol{h}_{\sigma}\|^2$} \Comment{\addblue{Round 2} with refined PSF}
        \For{$t = \tau:2$} \do \\
        \State $\hat{\z}_{t} \leftarrow \left(\z_t - \sqrt{1 - \bar\alpha_t} \boldsymbol{\mathcal{E}}_{\theta}^{(t)}(\z_t)  \right) / \sqrt{\bar\alpha_t}$
        \State $\hat{\X}_{t} \leftarrow \boldsymbol{D}_{\theta}(\hat{\z}_{t})$
        \State $\bar\X_t :=  \argmin_{\X \in \hat\X_t + \boldsymbol{\mathcal K}_l}\|\Y - \X \addblue{\ast \boldsymbol{h}_{\sigma}}\|^2$
        \State $\bar\X_{t} \leftarrow \bar\X_{t} \ast \boldsymbol{h}_{\sigma_t}$
        \State $\bar\z_t = \boldsymbol{E}_{\theta}(\bar\X_{t})$
        \State $\z_{t-1} = \sqrt{\bar\alpha_{t-1}}\bar\z_{t} + \sqrt{1 - \bar\alpha_{t-1}} \boldsymbol{\mathcal{E}}_{t}$ 
        \EndFor
        \State $\z_0 \leftarrow \left(\z_1 - \sqrt{1 - \bar\alpha_1} \boldsymbol{\mathcal{E}}_{\theta}^{(1)}(\z_1)  \right) / \sqrt{\bar\alpha_1}$
        \State \textbf{return} $\z_0$
   \end{algorithmic}
\end{algorithm}

In the context of blind deconvolution, an intuitive strategy is to alternate between point spread function (PSF) estimation and deconvolution. Since accurately estimating the initial PSF is challenging, we first employ a lightweight video deblurring module, DeepDeblur~\citep{nah2017deep}, for preliminary restoration. The initial PSF is then estimated based on this pre-restored video.
Using the estimated PSF, we perform a Round 1 reconstruction with our proposed method. Subsequently, the PSF is refined based on the output of this reconstruction. The refined PSF is then utilized for the final (Round 2) reconstruction, yielding an improved result.

\begin{figure*}[h]
  \centering
    \centerline{{\includegraphics[width=0.9\linewidth]{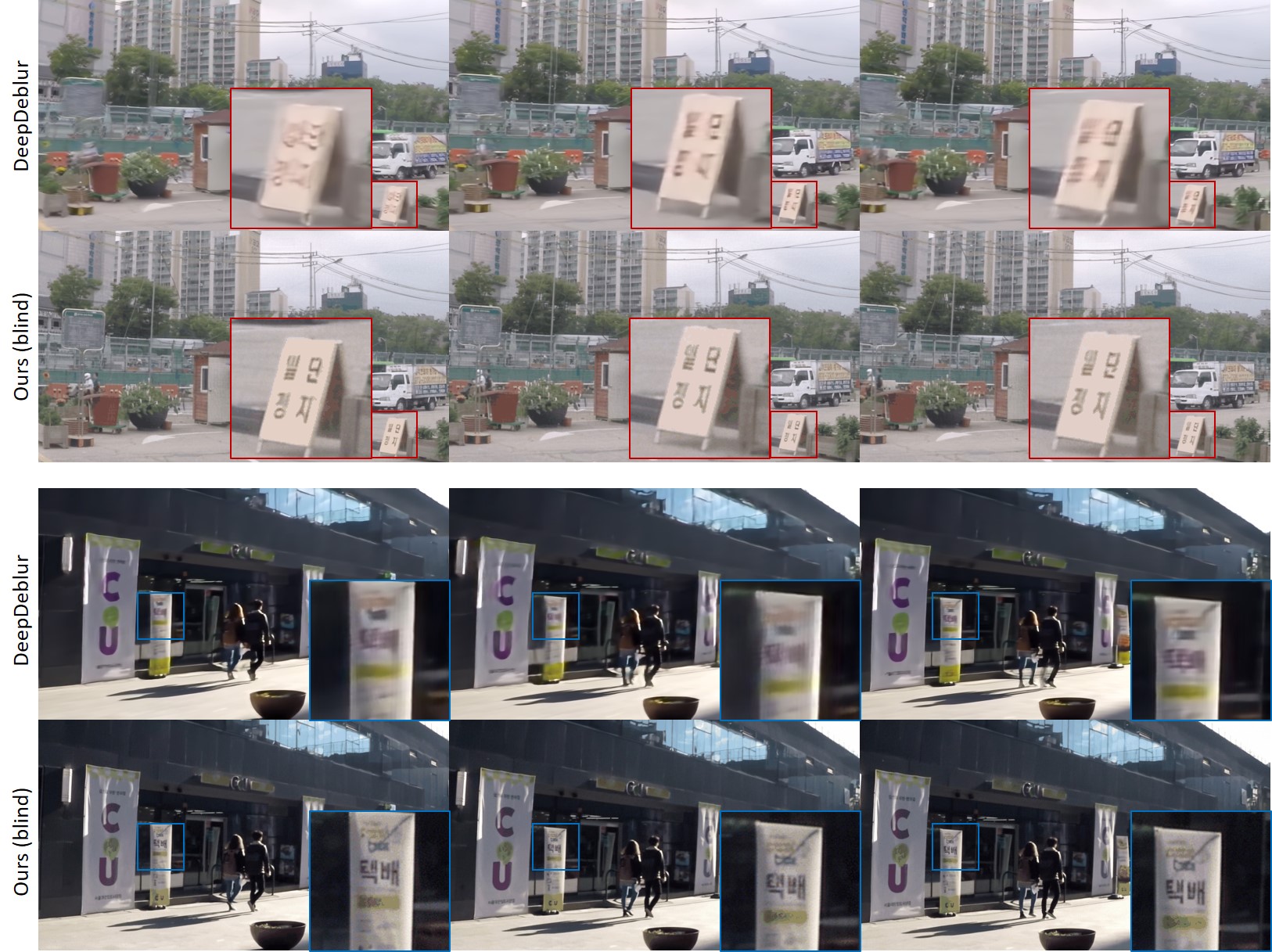}}}
    \caption{Qualitative comparison of video deblurring results on the GoPro test dataset~\cite{nah2017deep} compared with DeepDeblur~\cite{nah2017deep}.}
    \label{fig:result_blind}
\end{figure*}

In summary, our method incorporates a lightweight pre-restoration step to estimate the initial PSF and employs a two-round reconstruction pipeline to achieve high-quality restoration through PSF refinement. The detailed steps of the algorithm are outlined in Algorithm~\ref{alg:ours_blind}.

The GoPro dataset consists of 240 fps videos captured with a GoPro camera, where motion blur is synthetically generated by averaging 7 to 13 consecutive frames~\citep{nah2017deep}.
For our experiments, we used the GoPro test dataset and performed blind video reconstruction using Algorithm~\ref{alg:ours_blind}, generating blurred inputs by randomly averaging 7 to 13 frames. To evaluate the effectiveness of our approach, we compared our reconstruction results with those from the pre-restoration module.
Our method significantly improves reconstruction quality, yielding highly detailed results. As shown in Fig.~\ref{fig:result_blind}, zoomed-in views of signboards and billboards reveal that our method recovers fine details, such as text, with greater precision. This improvement demonstrates how incorporating a diffusion prior enables more accurate PSF estimation. Additionally, it highlights the potential of our method to extend to various blind inverse problems.

\section{Comprehensive visualizations}
For an in-depth understanding of the experimental results, we provide video visualizations on our anonymous project page\footnote{\url{https://vision-xl.github.io/}}.
The page features 36 paired visualizations of measurements and reconstructions across various aspect ratios and degradation types.
As shown on the project page, our method delivers highly satisfactory reconstruction results for various spatio-temporal inverse problems.

Additional comparisons with baselines are available on our supplementary anonymous project page\footnote{\url{https://vision-xl.github.io/supple/}}. 
In baseline comparisons, ADMM-TV struggles to reconstruct temporal degradations, and SVI~\cite{kwon2024solving} exhibits poor temporal consistency. 
DiffIR2VR~\cite{yeh2024diffir2vr} frequently fails to reconstruct and produces undesired artifacts, likely due to errors in the optical flow estimation module.
In contrast, our approach achieves superior performance across various spatio-temporal inverse problems.

We also provide visualizations of ablation studies. Regarding initialization effects, our pseudo-batch inversion significantly improves temporal consistency compared to random noise initialization or batch-consistent noise initialization~\cite{kwon2024solving}. 
Regarding the low-pass filter effect, we observe that applying a well-scheduled low-pass filter produces cleaner results with fewer artifacts. 
Without the low-pass filter, artifacts such as the grid pattern under the red bridge or the lattice-like texture on the body of the sea snake are noticeable.

We strongly encourage you to visit these project pages to explore the superior reconstruction performance of our method.

\end{document}